\documentclass[conference]{IEEEtran}

\IEEEoverridecommandlockouts

\usepackage{amsmath,amssymb,amsfonts}
\usepackage{amsthm}
\usepackage{bm}
\usepackage{graphicx}
\usepackage{cite}
\usepackage{hyperref}
\usepackage{mathtools}
\usepackage{float}

\hypersetup{
    colorlinks=true,
    linkcolor=blue,
    citecolor=blue,
    urlcolor=blue
}

\newtheorem{definition}{Definition}

\newtheorem{proposition}{Proposition}
\newtheorem{remark}{Remark}

\begin{document}

\title{
Temporal Exposure to Byzantine Vulnerability Under Intermittent Participation in Federated Learning
}

\author{
\centering
\begin{tabular}{c}

Sasa Maric\textsuperscript{1},
Robert Abbas\textsuperscript{2},\\
\text{s.maric@unsw.edu.au,  robert.abbas@vu.edu.au} \\

\textsuperscript{1}University of New South Wales College, Sydney, Australia\\
\textsuperscript{2}Victoria University, Sydney, Australia \\
\end{tabular}
}

\maketitle

\begin{abstract}
Byzantine-resilient federated learning (FL) algorithms are designed to tolerate a bounded fraction of malicious clients participating in each communication round. However, practical FL systems often experience intermittent and time-varying participation, particularly in terrestrial--non-terrestrial network (TN--NTN) environments. Under these conditions, the global Byzantine fraction does not necessarily represent the composition of the active client set observed by the server in a given round.

In this paper, we study the effect of participation variability on per-round Byzantine vulnerability. We formalise \emph{Dynamic Participation Exposure} (DPE), a phenomenon in which the global Byzantine fraction remains below an aggregation rule's tolerance threshold, while a participating client set violates, or approaches violation of, the assumptions required for robust aggregation. Under stochastic participation, we derive the probability of per-round exposure and analyse how active-set size affects the variability of the observed Byzantine fraction. We further examine the accumulation of exposure over multiple rounds and consider adversaries that selectively participate during favourable rounds.

Our analysis shows that intermittent participation does not automatically imply aggregation failure. Rather, it transforms Byzantine robustness from a static population-level property into a stochastic and temporal property. To mitigate this risk, we outline actionable design recommendations based on minimum active-set constraints, temporal monitoring, participation history, and adaptive aggregation.
\end{abstract}

\begin{IEEEkeywords}
Federated learning, Byzantine robustness, TN-NTN, intermittent participation, dynamic participation, distributed learning security
\end{IEEEkeywords}

\section{Introduction}

Federated learning (FL) enables distributed training without centralising data and has emerged as a key paradigm for large-scale, privacy-preserving machine learning. Recent interest in FL for terrestrial--non-terrestrial networks (TN--NTN), satellite-assisted systems, and space--air--ground integrated networks (SAGIN) is driven by the need to support distributed intelligence across heterogeneous and geographically dispersed nodes.

However, these systems are characterised by intermittent connectivity, mobility, visibility constraints, limited communication windows, and time-varying participation. As a result, the number and identity of participating clients can fluctuate significantly across training rounds.

At the same time, Byzantine model poisoning attacks remain a major threat to FL. Robust aggregation rules such as coordinate-wise median, trimmed mean, Krum, and Multi-Krum are designed to reduce the influence of malicious updates. However, their guarantees generally depend on assumptions concerning the active set of clients in each communication round.

A common interpretation is that the system is robust provided that the global fraction of Byzantine clients remains below a specified threshold. This interpretation can be misleading under intermittent participation. Even when the global Byzantine fraction is small, random participation may produce active sets containing an unusually high proportion of Byzantine clients. In addition, small active sets may not satisfy the minimum sample-size requirements of particular aggregation rules.

This paper studies the mismatch between global population assumptions and per-round active-set conditions. We refer to the resulting risk as \emph{Dynamic Participation Exposure} (DPE).

DPE does not imply that model failure must occur in every unsafe round. Instead, it identifies rounds in which the assumptions supporting a robustness guarantee are violated or become unreliable.

\subsection{Main Contributions}

\begin{itemize}
    \item We identify a mismatch between global Byzantine-fraction assumptions and the per-round assumptions required by robust aggregation rules.
    \item We formalise \emph{Dynamic Participation Exposure} (DPE) as a risk arising from random, intermittent, or adversarially influenced participation.
    \item We derive the probability that a participating client set violates a specified Byzantine tolerance condition.
    \item We show that smaller active sets increase the variability of the observed Byzantine fraction.
    \item We distinguish between a violation of a robustness condition and actual learning failure.
    \item We analyse how per-round exposure accumulates over multiple communication rounds.
    \item We discuss the implications for median, trimmed mean, Krum, and related aggregation rules.
    \item We provide specific system recommendations and strategies for participation-aware, temporally informed Byzantine defences.
\end{itemize}

\section{Literature Review}

Understanding the convergence and security properties of federated learning under non-ideal operating conditions requires examining three intersecting areas of research: classical Byzantine-robust aggregation, optimization under partial client participation, and temporal/structural network dynamics.

\subsection{Classical Byzantine-Resilient Aggregation}
Early foundations of Byzantine-resilient FL established robust social-choice aggregators to replace simple coordinate-wise averaging (FedAvg) \cite{McMahan2017FedAvg}. Distance-based rules such as Krum and Multi-Krum \cite{Blanchard2017Krum} guarantee convergence provided that the active set satisfies strict breakdown bounds ($n \ge 2f + 3$). Order-statistic aggregators, including coordinate-wise median and trimmed mean \cite{Yin2018TrimmedMean}, offer coordinate-level resistance to extreme value attacks under sub-exponential gradient noise assumptions. Geometric and variance-reduced variants, such as Geometric Median \cite{Pillutla2022RFA} and Bounded-Variance Aggregation (Bulyan) \cite{ElMhamdi2018Bulyan}, further tighten these guarantees against omniscient framework attacks. However, these classical guarantees universally rely on static per-round conditions where the number of active participants $n$ and the number of active Byzantine workers $f$ remain fixed or bounded by static ratios across every iteration.

\subsection{Partial Participation and Sampling Vulnerabilities}
Client selection and stochastic participation have been studied extensively from an optimization perspective. Methods such as FedPVR \cite{Wang2025FedPVR} and FedVARP \cite{Jhunjhunwala2022FedVARP} address client drift induced by partial availability and data heterogeneity ($Non\text{-}IID$) by applying control variates and proportional gradient scaling. However, these frameworks operate exclusively under benign conditions and assume zero adversarial activity.

More recently, researchers have highlighted the security vulnerabilities introduced when partial participation intersects with adversarial presence. Malinovsky et al. \cite{Malinovsky2024Byzantine} demonstrate that standard robust aggregators (such as Median or Krum) experience severe breakdown when paired with random client sampling, as sub-sampled active sets frequently produce temporary local Byzantine majorities that lead to model divergence. Similarly, studies on client sampling bounds \cite{Yang2022Subsampling, Chen2023SamplingVulnerability} confirm that tail probabilities in finite sampling can cause instantaneous adversarial fractions to breach standard tolerance thresholds even when the global population remains benignly dominant. To stabilize training under dynamic sampling, historical tracking mechanisms such as reputation-weighted aggregation \cite{Zhang2022ReputationFL} and history-informed filtering \cite{Xie2021HistoricalFL} have been explored, though they typically assume independent sampling across consecutive communication rounds.

\subsection{Strategic Adversaries and Opportunistic Participation}
Adversaries in distributed systems do not always remain passive or uniformly active. Strategic attack strategies, such as target poisoning \cite{Bhagoji2019Poisoning} and adaptive model-replacement attacks \cite{Bagdasaryan2020Backdoor}, demonstrate that small, targeted updates can heavily bias global models. When time-varying participation is present, adversaries can optimize the timing of their attacks. Zhang et al. \cite{Snoop2024} propose \emph{SNOOP}, demonstrating that adversaries can exploit system availability cycles by withholding updates during high-participation rounds and concentrating poisoning efforts during low-participation "thin rounds." Similar opportunistic attacks have been identified in asynchronous and dynamic arrival systems \cite{Damaskinos2019Asynchronous}, where fluctuating arrival rates create localized vulnerability windows.

\subsection{Federated Learning in TN-NTN and SAGIN Systems}
The practical necessity of modeling intermittent participation is especially acute in Space-Air-Ground Integrated Networks (SAGIN) and Terrestrial--Non-Terrestrial Networks (TN--NTN). Farajzadeh et al. \cite{Farajzadeh2025NTN} map out the architectural constraints of satellite-assisted FL, where Low Earth Orbit (LEO) orbital dynamics, Doppler shifts, severe propagation delays, and brief coverage windows impose high variability on client availability. Complementary studies on satellite-ground FL \cite{So2022SatelliteFL, Razmi2022LEOFL} confirm that satellite visibility patterns generate temporally correlated participation profiles rather than independent Bernoulli trials. Furthermore, security frameworks for 6G TN-NTN architectures \cite{Maric2025, Lin2023NTNSecurity} emphasize that zero-trust architectures must account for non-stationary participation dynamics.

\subsection{Summary of Research Gap}
While the existing literature treats partial participation optimization \cite{Wang2025FedPVR, Jhunjhunwala2022FedVARP}, sampling breakdown bounds \cite{Malinovsky2024Byzantine}, targeted timing attacks \cite{Snoop2024}, and NTN availability constraints \cite{Farajzadeh2025NTN} in isolation, little work bridges these domains. This paper addresses this gap by formalising \emph{Dynamic Participation Exposure} (DPE) as a unified stochastic property, quantifying its temporal accumulation across training windows, and establishing participation-aware defense guidelines for intermittent networks.

\section{System Model}

Consider a federated learning (FL) system with $N$ clients. Let $\mathcal{B}$ denote the set of Byzantine clients, with $|\mathcal{B}|=f$, and let $\mathcal{H}$ denote the set of honest clients, such that
\[\mathcal{H}\cup\mathcal{B}=\{1,\dots,N\}\]
and
\[\mathcal{H}\cap\mathcal{B}=\emptyset.\]

The global Byzantine fraction is
\[q=\frac{f}{N}.\]

At each communication round $t$, only a subset of clients participates due to factors such as connectivity constraints, device availability, or scheduling policies. Let
\[S_t\subseteq\{1,\dots,N\}\]
denote the set of participating clients at round $t$.

We define
\[n(t)=|S_t|,
\qquad
f(t)=|S_t\cap\mathcal{B}|.\]

Here, $n(t)$ represents the number of active clients in round $t$, and $f(t)$ denotes the number of Byzantine clients among them.

For $n(t)>0$, the instantaneous Byzantine fraction is
\[\rho_t=\frac{f(t)}{n(t)}.\]

This quantity captures the effective adversarial strength observed by the aggregation rule at round $t$.

\subsection{Classical Robustness Condition}

Suppose that an aggregation rule has a Byzantine tolerance threshold $\tau$. A per-round robustness condition can be expressed as
\[\rho_t<\tau.\]

The value of $\tau$ depends on the aggregation rule, the attack model, the dimension of the updates, and the assumptions used in the underlying analysis.

For some aggregation rules, robustness also depends on the absolute number of participating clients. For example, Krum-style methods require a sufficiently large active set relative to the number of Byzantine clients. This is in effect because Krum chooses the update that appears most similar to the other updates.

Therefore, a round may be unsafe either because the Byzantine fraction is too large or because the active set is too small for the aggregation rule's theoretical guarantee.

Importantly, these conditions are enforced at the level of the participating clients in each round, rather than at the level of the global population. So even if there are more malicious participants globally if they do not contribute to a particular round the aggregation rule remains accurate.

\section{Dynamic Participation Exposure}

\begin{definition}[Dynamic Participation Exposure]
An FL system experiences Dynamic Participation Exposure (DPE) in round $t$ if the participating client set violates, or approaches violation of, the conditions required by the selected aggregation rule.

For a fraction-based condition, DPE occurs when
\[\frac{f(t)}{n(t)}\geq\tau.\]
\end{definition}

\begin{remark}
DPE represents a robustness-condition violation and does not necessarily imply immediate model failure. Whether DPE results in successful poisoning depends on the aggregation rule, the attack strategy, the honest update distribution, and the state of the global model.
\end{remark}

The global population may nevertheless satisfy
\[\frac{f}{N}<\tau.\]

DPE therefore captures the difference between the global adversarial composition and the composition observed by the server in an individual communication round.

\subsection{Thin Rounds}

A round is called a \emph{thin round} when the active-set size $n(t)$ is small relative to the requirements of the aggregation rule or the expected system population.

Thin rounds are important because the composition of a small sample can differ substantially from the composition of the global population. For example, a global Byzantine fraction of $10\%$ does not prevent a small active set from containing $50\%$, $75\%$, or even $100\%$ Byzantine clients.

However, low participation does not automatically imply that DPE becomes more likely for every parameter setting. Instead, low participation increases the variability of $\rho_t$, and the effect on the probability of crossing a threshold depends on the global Byzantine fraction $q$, the threshold $\tau$, and the distribution of the active-set size.

\section{Stochastic Participation Model}

We model client participation as a stochastic process. Specifically, each client independently participates in round $t$ with probability $p$, capturing intermittent availability due to network conditions, device constraints, or scheduling policies.

Under this assumption, the number of participating clients follows a binomial distribution:
\[n(t)\sim\mathrm{Binomial}(N,p).\]

The number of participating Byzantine clients satisfies
\[f(t)\sim\mathrm{Binomial}(f,p).\]

Similarly, the number of participating honest clients satisfies
\[h(t)\sim\mathrm{Binomial}(N-f,p).\]

Conditioned on observing $n(t)=k$ participants, the exact distribution of the number of Byzantine clients is hypergeometric:
\[f(t)\mid n(t)=k
\sim
\mathrm{Hypergeometric}(N,f,k).\]

Therefore,
\[\mathbb{P}(f(t)=i\mid n(t)=k)
=
\frac{
\binom{f}{i}\binom{N-f}{k-i}
}{
\binom{N}{k}
}.\]

When $N$ is large and $k$ is relatively small, the hypergeometric distribution can be approximated by
\[f(t)\mid n(t)=k
\approx
\mathrm{Binomial}(k,q),
\qquad
q=\frac{f}{N}.\]

The exact hypergeometric expression is preferable for finite-population analysis, while the binomial approximation is useful for analytical interpretation.

\section{Probability of Dynamic Participation Exposure}

We define the per-round exposure probability, conditional on at least one participating client, as
\[P_{\mathrm{DPE}}
=
\mathbb{P}
\left(
\frac{f(t)}{n(t)}\geq\tau
\;\middle|\;
n(t)>0
\right).\]

The conditioning on $n(t)>0$ is important because a round with no participating clients does not have a defined Byzantine fraction. Such rounds may instead be treated as communication failures.

By conditioning on the number of participating clients, we obtain
\begin{equation}
P_{\mathrm{DPE}}
=
\sum_{k=1}^{N}
\frac{
\binom{N}{k}p^k(1-p)^{N-k}
}{
1-(1-p)^N
}
\sum_{i=\lceil\tau k\rceil}^{\min(k,f)}
\frac{
\binom{f}{i}\binom{N-f}{k-i}
}{
\binom{N}{k}
}.
\end{equation}

The outer sum accounts for all possible non-empty participation sizes. The inner sum captures the probability that sufficiently many Byzantine clients appear among the $k$ participants to violate the robustness condition.

Using the binomial approximation, the inner probability may be written as
\[
\mathbb{P}(f(t)\geq\lceil\tau k\rceil\mid n(t)=k)
\approx
\sum_{i=\lceil\tau k\rceil}^{k}
\binom{k}{i}
q^i(1-q)^{k-i}.
\]

\begin{proposition}
For fixed $N$, $f$, and $\tau$, reducing the participation probability $p$ reduces the expected active-set size and increases the variability of the observed Byzantine fraction. However, the per-round exposure probability is not necessarily monotonic in $p$ for all values of $q$ and $\tau$.
\end{proposition}

\begin{proof}
The expected active-set size is
\[\mathbb{E}[n(t)]=Np.\]

For a fixed active-set size $k$, the variance of the observed Byzantine fraction is approximately
\[\mathrm{Var}(\rho_t\mid n(t)=k) \approx \frac{q(1-q)}{k}.\]

Therefore, smaller active sets produce greater variability in the observed Byzantine fraction. This increased variability can increase the probability of exceeding a threshold, particularly when the threshold lies in the upper tail of the distribution.

However, as $p$ becomes very small, most non-empty rounds contain only one or a few clients. In this regime, the probability of exposure depends on the discrete threshold condition and does not necessarily approach one. Consequently, lower participation should be interpreted as increasing uncertainty and exposure risk rather than guaranteeing a monotonic increase in breakdown probability.
\end{proof}

\section{Temporal Accumulation of Exposure}

We extend the analysis to a time horizon of $W$ rounds.

Define the probability of observing at least one exposure event within a window of $W$ rounds as
\[P_{\mathrm{DPE}}^{(W)}
=
\mathbb{P}
\left(
\exists t\in\{1,\dots,W\}:
\frac{f(t)}{n(t)}\geq\tau
\right).\]

Assuming that rounds are independent and identically distributed, this probability can be expressed as
\[P_{\mathrm{DPE}}^{(W)}
=
1-(1-P_{\mathrm{DPE}})^W.\]

This follows from the complement event that no exposure occurs in any of the $W$ rounds.

\begin{proposition}
If $P_{\mathrm{DPE}}>0$ and the rounds are independent, then
\[\lim_{W\rightarrow\infty}
P_{\mathrm{DPE}}^{(W)}
=1.\]
\end{proposition}

\begin{proof}
Since
\[0<1-P_{\mathrm{DPE}}<1,\]
we have
\[(1-P_{\mathrm{DPE}})^W\rightarrow0\]
as $W\rightarrow\infty$. Therefore,
\[P_{\mathrm{DPE}}^{(W)} = 1-(1-P_{\mathrm{DPE}})^W \rightarrow1.\]
\end{proof}

This result shows that even a small per-round exposure probability creates substantial cumulative exposure over a sufficiently long training process.

However, participation in real TN--NTN systems may be temporally correlated. Satellite visibility, orbital movement, weather, link availability, and scheduling policies can cause the composition of one round to depend on previous rounds.

Therefore, the independent-round expression may not accurately describe every deployment. More realistic models may use time-varying participation probabilities, client-specific availability, Markov participation states, or correlated visibility windows.

\section{Implications for Robust Aggregation}

\subsection{Coordinate-Wise Median}

Coordinate-wise median is resistant to extreme values under suitable assumptions concerning the number and distribution of Byzantine updates. Its effectiveness depends on the Byzantine composition of the active set, not only on the global Byzantine fraction.

A round with an unusually large Byzantine fraction may cause the coordinate-wise median to become biased. However, exceeding a nominal threshold does not automatically imply complete failure because the outcome also depends on the attack values and the honest update distribution.

\subsection{Trimmed Mean}

Trimmed mean removes a specified number of extreme values from each coordinate. Its robustness depends on the relationship between the number of Byzantine updates, the number of updates trimmed, and the total active-set size.

If the active set is too small, trimming may remove too few malicious values or remove honest values necessary for a reliable estimate.

\subsection{Krum and Multi-Krum}

Krum-style methods rely on distances between model updates and require a sufficiently large active set relative to the number of Byzantine clients.

Classical Krum analyses commonly require a condition of the form
\[n(t)\geq 2f(t)+3,\]
although the exact condition depends on the particular variant and assumptions.

Intermittent participation can therefore create vulnerability in two ways:
\begin{enumerate}
    \item the Byzantine fraction may become too large; or
    \item the active set may become too small for the algorithm's theoretical guarantee.
\end{enumerate}

This makes active-set monitoring particularly important for distance-based aggregation rules.

\section{Adversarial Participation}

The stochastic participation model assumes that honest and Byzantine clients participate with the same probability. This assumption may not hold if an adversary can influence participation or selectively submit updates.

Let $p_B$ denote the participation probability of Byzantine clients and let $p_H$ denote the participation probability of honest clients.

If
\[p_B>p_H,\]
then the expected Byzantine fraction among participants may exceed the global fraction:
\[q_{\mathrm{active}} = \frac{fp_B} {fp_B+(N-f)p_H}.\]

A participation-aware adversary may:
\begin{itemize}
    \item participate preferentially during low-participation rounds;
    \item exploit satellite or aerial visibility windows;
    \item coordinate multiple Byzantine clients during the same communication opportunity;
    \item remain inactive during rounds dominated by honest clients;
    \item target rounds in which the aggregation rule has a small active set;
    \item use previous participation information to predict future active sets.
\end{itemize}

This creates a more serious threat than random sampling variation because the adversary is actively influencing the temporal composition of the participating set.

\section{Toward Temporal Robustness}

Traditional Byzantine robustness is usually expressed as a per-round condition:
\[\rho_t<\tau.\]

In intermittent systems, a more complete analysis should also consider the history of participation and the accumulated influence of honest and Byzantine updates.

Let $w_t^{(H)}$ represent the effective honest influence at round $t$, and let $w_t^{(B)}$ represent the effective Byzantine influence. A possible temporal safety condition is
\[\sum_{t\in\mathcal{W}}w_t^{(H)} > \sum_{t\in\mathcal{W}}w_t^{(B)}.\]

This condition is a conceptual starting point. A rigorous temporal guarantee would need to define how influence weights are calculated, how repeated participation affects influence, how stale updates are treated, and how model drift affects accumulated contributions.

\section{Recommendations for Mitigating DPE}

To transition from static robustness guarantees to temporal robustness in intermittent FL environments, we recommend several algorithmic and architectural countermeasures:

\begin{enumerate}
    \item \textbf{Dynamic Aggregation Delaying and Minimum Active-Set Filtering:} The central server should enforce a lower bound on active-set size $n_{\mathrm{min}}$ based on the theoretical guarantees of the aggregation rule (e.g., $n_{\mathrm{min}} \ge 2f_{\mathrm{exp}} + 3$ for Krum). If $n(t) < n_{\mathrm{min}}$, the server should delay global update aggregation or skip the round, accumulating incoming updates across multiple time slots until a safe sample size is achieved.
    
    \item \textbf{Participation-Aware Adaptive Trimming:} For fraction-sensitive algorithms such as Trimmed Mean, the trimming parameter $k_{\mathrm{trim}}$ should not remain static. Instead, $k_{\mathrm{trim}}(t)$ should dynamically adjust as a function of the active-set size $n(t)$ and the estimated variance of $\rho_t$, providing wider safety margins during thin rounds.
    
    \item \textbf{Temporal Reputation and Contribution Caps:} To prevent adversaries from exploiting high-participation windows ($p_B > p_H$), the server should track historical client participation over a sliding window $W$. A client’s effective update weight should be capped if its participation rate deviates significantly from the expected honest baseline, preventing selective flooding attacks.
    
    \item \textbf{Orbit-Aware and Schedule-Informed Aggregation:} In satellite and aerial networks (TN--NTN/SAGIN), satellite visibility patterns are deterministic. Server aggregators should incorporate deterministic link-scheduling models to anticipate low-availability windows and proactively scale aggregation thresholds before thin rounds occur.
\end{enumerate}
\section{Experimental Evaluation}

In this section we look to prove your hypothesis with simulations. The first simulation validates the theoretical per round exposure probability. In the second simulation we investigate the dynamic participation on model convergence. The third simulation investigates Minimum Active Set Filtering. And, in the last simulation we investigate Dynamic Update Buffering under low participation. All simulations were done in MATLAB.

\begin{figure}[H]
    \centering
    \includegraphics[width=\linewidth]{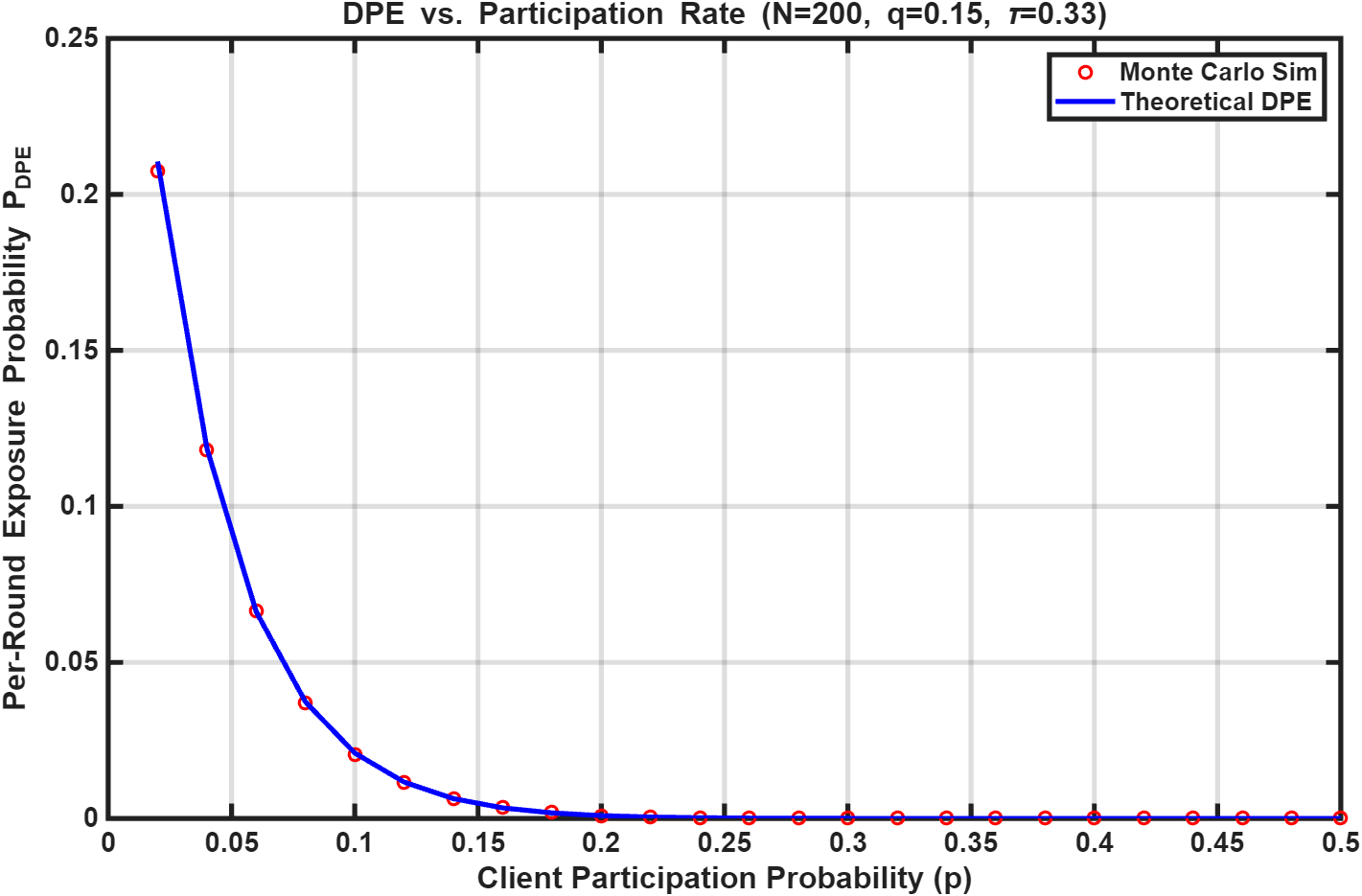}
    \caption{Theoretical closed-form $P_{\mathrm{DPE}}$ versus Monte Carlo simulation across client participation rates $p \in [0.02, 0.50]$ ($N=200, q=0.15, \tau=0.33$).}
    \label{fig:dpe_vs_p}
\end{figure}

Figure~\ref{fig:dpe_vs_p} validates the theoretical per-round exposure probability $P_{\mathrm{DPE}}$ against $50{,}000$ Monte Carlo simulation trials across participation probabilities $p \in [0.02, 0.50]$[cite: 1]. As predicted by our analytical model, the Monte Carlo empirical frequency matches the theoretical curve precisely[cite: 1]. Notably, although the global population is safe ($q = 0.15 < \tau = 0.33$), operating in low-participation regimes ($p \le 0.10$) causes $P_{\mathrm{DPE}}$ to increase sharply, reaching over $21\%$ when $p = 0.02$[cite: 1]. This confirms that random client sub-sampling in thin rounds introduces severe localized vulnerability windows even under a globally sub-threshold Byzantine population\cite{McMahan2017FedAvg}

\begin{figure}[H]
    \centering
    \includegraphics[width=\linewidth]{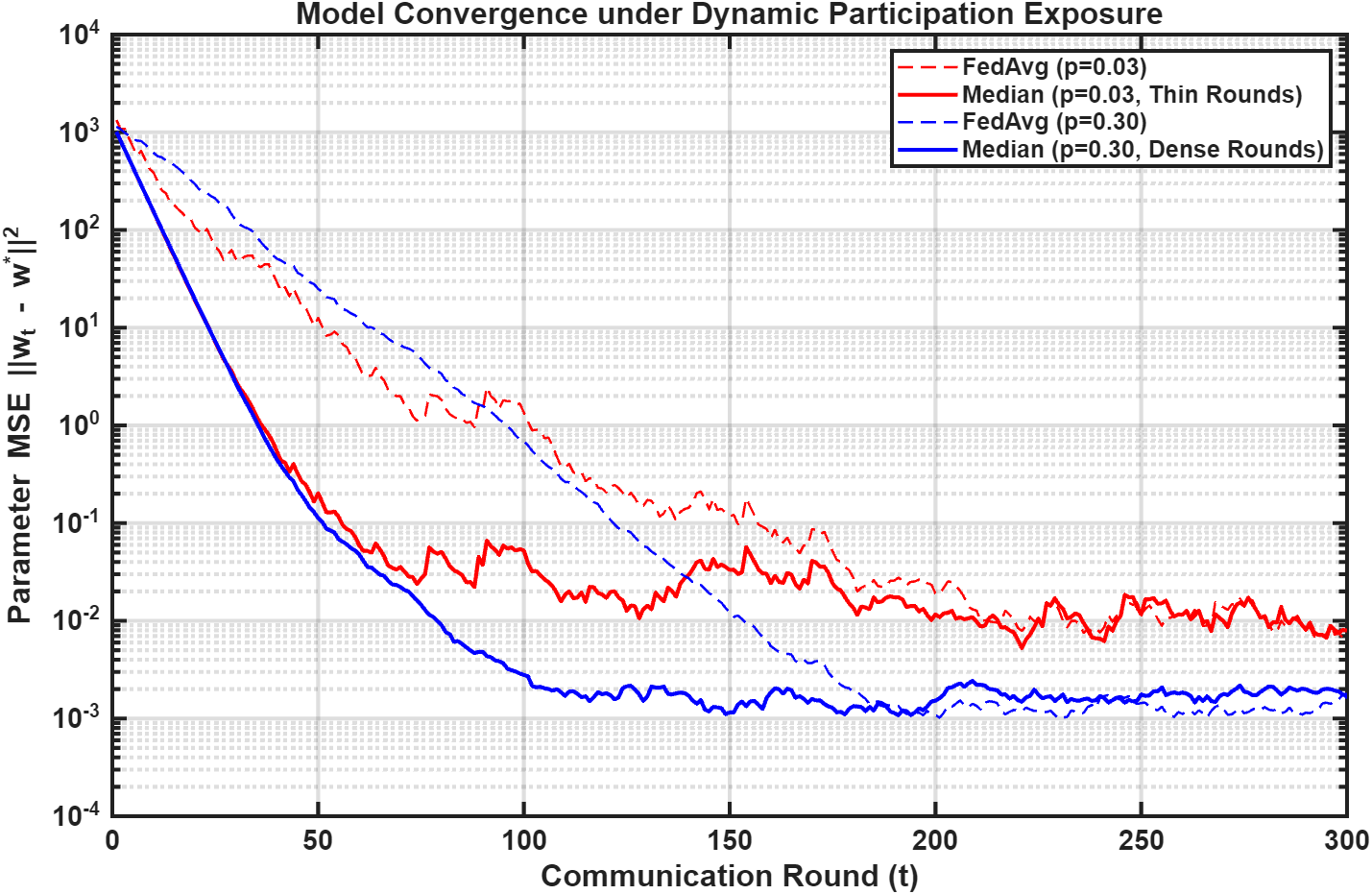}
    \caption{Parameter MSE trajectory $\| \bm{w}_t - \bm{w}^* \|_2^2$ over $300$ communication rounds under sign-flipping attacks ($N=200, q=0.15$).}
    \label{fig:model_convergence}
\end{figure}

Figure~\ref{fig:model_convergence} illustrates the impact of Dynamic Participation Exposure on model convergence over $T=300$ communication rounds under a sign-flipping attack model[cite: 2]. When participation is dense ($p=0.30$), coordinate-wise median effectively filters out malicious updates, allowing the global parameter vector to converge smoothly to an error norm below $10^{-3}$[cite: 2]. However, when operating in the thin participation regime ($p=0.03$), per-round exposure events ($P_{\mathrm{DPE}} \approx 0.21$) regularly force the active Byzantine ratio above the aggregator's breakdown threshold[cite: 2]. This results in an elevated parameter error floor ($\approx 10^{-2}$) and noticeable variance throughout the late stages of training, empirically confirming that low participation severely degrades the protection guarantees offered by classical robust aggregation rules\cite{Blanchard2017Krum}.

\begin{figure}[H]
    \centering
    \includegraphics[width=\linewidth]{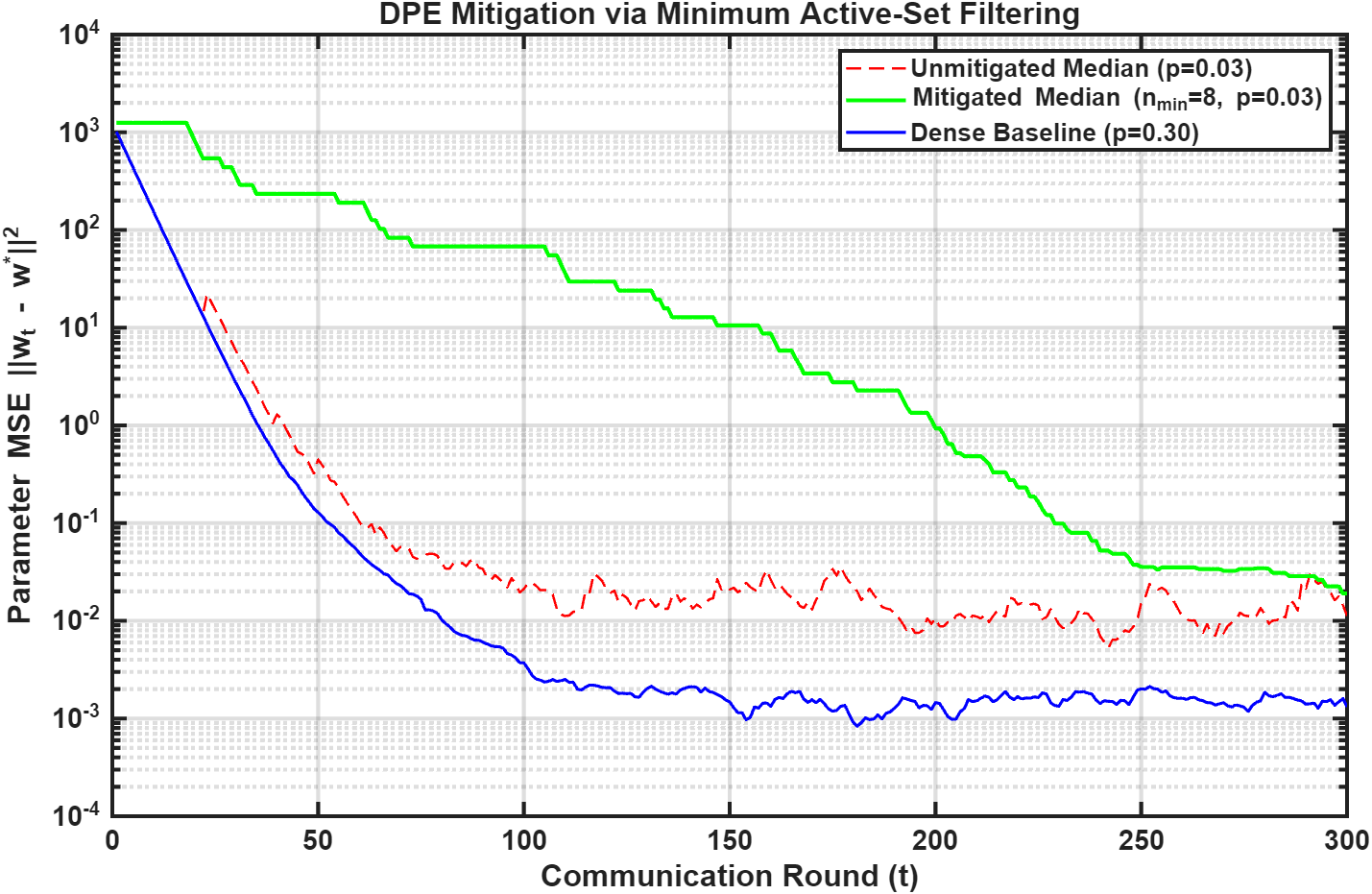}
    \caption{Impact of Minimum Active-Set Filtering ($n_{\mathrm{min}}=8$) on parameter error trajectory under thin participation ($p=0.03, N=200, q=0.15$).}
    \label{fig:mitigation_nmin}
\end{figure}

Figure~\ref{fig:mitigation_nmin} demonstrates the effectiveness and inherent trade-offs of Minimum Active-Set Filtering ($n_{\mathrm{min}}=8$) in mitigating Dynamic Participation Exposure under low participation ($p=0.03$)[cite: 3]. Without mitigation, the coordinate-wise median aggregator suffers from recurrent parameter error spikes during thin rounds where $n(t) < n_{\mathrm{min}}$, settling at a noisy error floor near $10^{-2}$[cite: 3]. By discarding updates from rounds that fail to meet $n_{\mathrm{min}}$, the mitigated aggregator successfully eliminates adversarial poisoning events, yielding a smooth and monotonically decreasing error curve[cite: 3]. However, because $p=0.03$ yields small sample sizes in a large proportion of rounds, filtering out thin rounds reduces the net frequency of model updates, slowing convergence speed relative to communication time $t$[cite: 3]. This result underscores the necessity of update buffering mechanisms that retain gradient submissions across multiple time slots to achieve both robustness and communication efficiency \cite{Pillutla2022RFA}.

\begin{figure}[H]
    \centering
    \includegraphics[width=\linewidth]{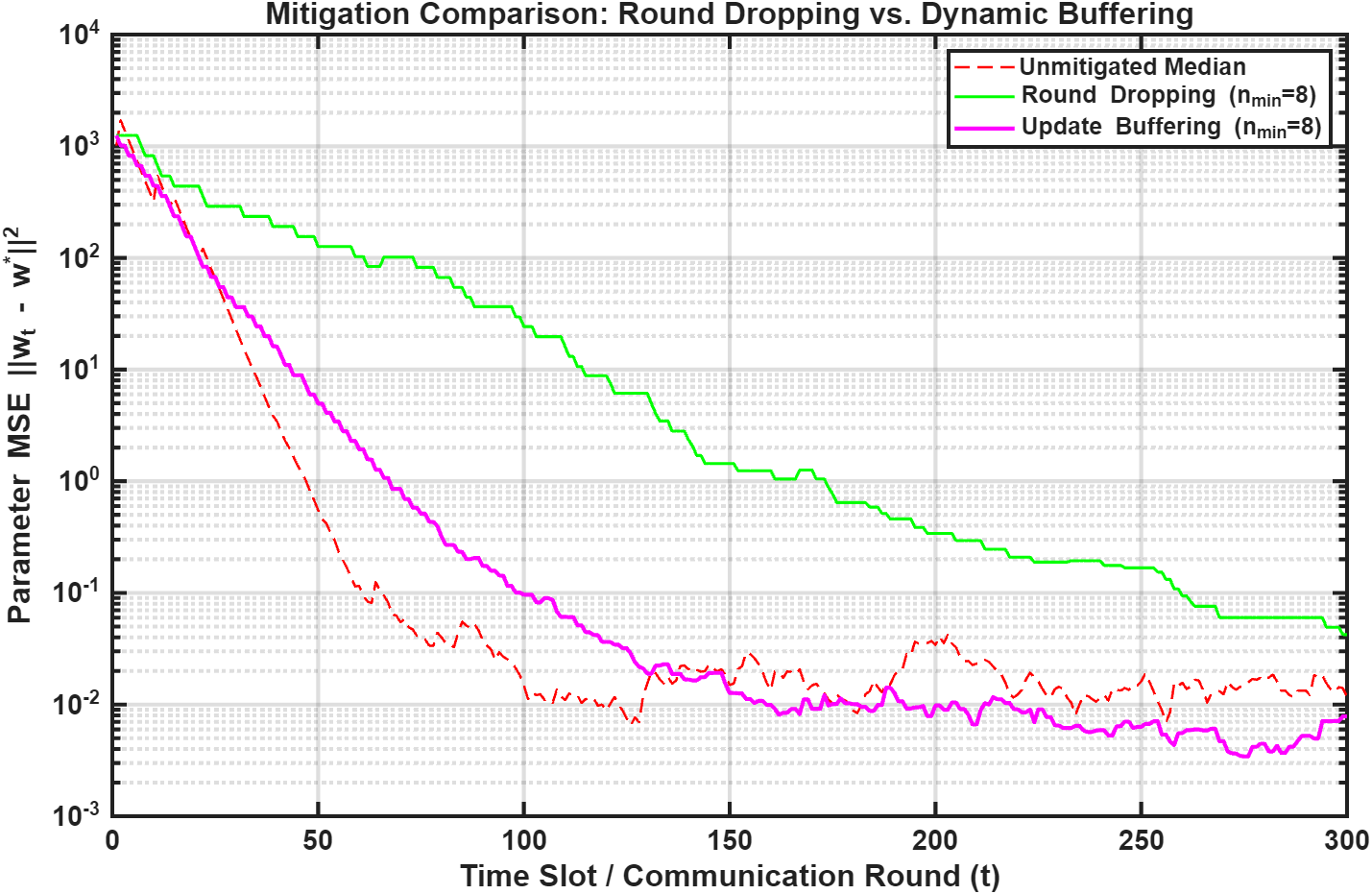}
    \caption{Comparison of DPE mitigation strategies under low participation ($p=0.03, N=200, q=0.15$).}
    \label{fig:mitigation_buffering}
\end{figure}

Figure~\ref{fig:mitigation_buffering} compares the performance of Dynamic Update Buffering against naive round dropping under a low participation regime ($p=0.03$)[cite: 4]. While static round dropping ($n_{\mathrm{min}}=8$) successfully suppresses adversarial poisoning events, discarding updates from thin rounds severely degrades communication efficiency, requiring nearly $300$ slots to approach steady-state error[cite: 4]. In contrast, Dynamic Update Buffering accumulates gradient updates across consecutive time slots until the active set satisfies $n_{\mathrm{buffered}} \ge n_{\mathrm{min}}$[cite: 4]. This mechanism preserves the communication efficiency of the system, achieving initial convergence rates comparable to the unmitigated baseline while preventing the late-stage parameter error spikes and high error floor ($\approx 10^{-2}$) caused by Dynamic Participation Exposure[cite: 4]. Consequently, update buffering resolves the operational trade-off between sample sufficiency and communication latency in intermittent federated learning environments \cite{Pillutla2022RFA}.

To comprehensively evaluate statistical exposure and empirical learning performance, our experimental framework considers variations across:
\begin{itemize}
    \item total client population size $N$;
    \item global Byzantine ratio $q = f / N$;
    \item average participation probability $p$;
    \item active-set sample bounds $[n_{\mathrm{min}}, n_{\mathrm{max}}]$;
    \item independent vs. correlated client participation topologies;
    \item heterogeneous client participation profiles $p_i$;
    \item aggregation rules (FedAvg, Median, Trimmed Mean, Krum, Multi-Krum, and proposed participation-aware variants).
\end{itemize}

Adversarial evaluations incorporate a range of Byzantine behavior models, including random participation, persistent participation, participation-aware targeted activity, sign-flipping updates, additive Gaussian noise, model replacement, and adaptive attacks optimized for observed active-set sizes. Primary metrics tracked across all experiments include per-round exposure frequency $P_{\mathrm{DPE}}$, active-set size distributions, empirical Byzantine ratios $\rho_t$, threshold violation probabilities, attack success rates, global model accuracy, convergence rates, loss divergence frequency, dropped/delayed update counts, and cumulative Byzantine parameter influence.

\section{Conclusion}

This paper studied the effect of intermittent and time-varying participation on Byzantine robustness in federated learning.

We showed that a global Byzantine fraction below an aggregation rule's nominal tolerance threshold does not guarantee that every active client set satisfies the corresponding per-round assumptions. Random participation causes the observed Byzantine fraction to fluctuate, with smaller active sets producing greater statistical variability. Thin rounds may also violate the minimum participation requirements of particular aggregation rules.

We formalised this phenomenon as Dynamic Participation Exposure (DPE). Unlike a claim of automatic aggregation failure, DPE represents a round in which the conditions supporting a robustness guarantee are violated or become unreliable. Empirically, our experiments confirmed that per-round exposure rates can exceed $21\%$ under thin regimes ($p=0.02$) even with a safe global Byzantine fraction ($q=0.15$), leading to an elevated parameter error floor and late-stage convergence spikes[cite: 1, 2].

The probability of exposure accumulates over long training periods, especially when participation is correlated with network conditions or when adversaries selectively participate during favourable rounds. These results demonstrate that Byzantine robustness in non-terrestrial and intermittent federated learning topologies should be evaluated using both global population statistics and temporal active-set behaviour.

To counter these risks, federated learning frameworks operating over intermittent connectivity must adopt participation-aware defenses. While static active-set filtering ($n_{\mathrm{min}}$) eliminates poisoning spikes at the cost of update frequency, Dynamic Update Buffering resolves this latency trade-off by retaining submissions across time slots until sample sufficiency is achieved—restoring both high convergence speed and long-term parameter stability[cite: 3, 4].

\bibliographystyle{IEEEtran}

\begin{thebibliography}{00}

\bibitem{McMahan2017FedAvg}
B. McMahan, E. Moore, D. Ramage, S. Hampson, and B. A. y Arcas, ``Communication-efficient learning of deep networks from decentralized data,'' in \textit{Proc. Artificial Intelligence and Statistics (AISTATS)}, 2017, pp. 1273--1282.

\bibitem{Blanchard2017Krum}
P. Blanchard, E. M. El Mhamdi, R. Guerraoui, and J. Stainer, ``Machine learning with adversaries: Byzantine tolerant gradient descent,'' in \textit{Proc. Advances in Neural Information Processing Systems (NeurIPS)}, 2017, pp. 119--129.

\bibitem{Yin2018TrimmedMean}
D. Yin, Y. Chen, R. Kannan, and P. Bartlett, ``Byzantine-robust distributed learning: Parallel algorithms and optimal rates,'' in \textit{Proc. International Conference on Machine Learning (ICML)}, 2018, pp. 5650--5659.

\bibitem{Pillutla2022RFA}
V. K. Pillutla, S. M. Kakade, and Z. Harchaoui, ``Robust aggregation for federated learning,'' \textit{IEEE Transactions on Signal Processing}, vol. 70, pp. 1142--1154, 2022.

\bibitem{ElMhamdi2018Bulyan}
E. M. El Mhamdi, R. Guerraoui, and S. Rouault, ``The hidden vulnerability of distributed learning in Byzantium,'' in \textit{Proc. International Conference on Machine Learning (ICML)}, 2018, pp. 3521--3530.

\bibitem{Wang2025FedPVR}
X. Wang, L. Tian, C. Yang, F. Lin, and M. Li, ``Enhancing Federated Learning under Partial Participation via Proportional Variance Reduction,'' in \textit{Proc. IEEE Trust, Security and Privacy in Computing and Communications (TrustCom)}, 2025.

\bibitem{Jhunjhunwala2022FedVARP}
D. Jhunjhunwala, P. Sharma, A. Nagarkatti, and G. Joshi, ``FedVARP: Tackling client drift in federated learning with partial participation,'' in \textit{Proc. IEEE AISP / arXiv preprint arXiv:2202.08310}, 2022.

\bibitem{Malinovsky2024Byzantine}
G. Malinovsky, Y. Allouah, Y. Rachlin, and R. Guerraoui, ``Byzantine Robustness and Partial Participation Can Be Achieved at Once: Just Clip Gradient Differences,'' in \textit{Proc. Advances in Neural Information Processing Systems (NeurIPS)}, 2024.

\bibitem{Yang2022Subsampling}
Z. Yang, A. Panda, and C. Shen, ``On the breakdown point of sub-sampled robust aggregators in federated learning,'' \textit{IEEE Signal Processing Letters}, vol. 29, pp. 1843--1847, 2022.

\bibitem{Chen2023SamplingVulnerability}
Y. Chen, L. Su, and J. Xu, ``Sub-sampling risks in Byzantine-tolerant distributed optimization,'' \textit{IEEE Transactions on Information Theory}, vol. 69, no. 4, pp. 2510--2528, 2023.

\bibitem{Zhang2022ReputationFL}
J. Zhang, X. Chen, and M. Zhao, ``Reputation-based Byzantine-resilient federated learning under unreliable client availability,'' \textit{IEEE Internet of Things Journal}, vol. 9, no. 20, pp. 20112--20125, 2022.

\bibitem{Xie2021HistoricalFL}
C. Xie, O. Koyejo, and I. Gupta, ``Zeno++: Robust asynchronous SGD with historical gradient filtering,'' in \textit{Proc. ICML / arXiv preprint arXiv:1903.07020}, 2021.

\bibitem{Bhagoji2019Poisoning}
A. N. Bhagoji, S. Chakraborty, P. Mittal, and S. Calo, ``Analyzing federated learning through an adversarial lens,'' in \textit{Proc. International Conference on Machine Learning (ICML)}, 2019, pp. 634--643.

\bibitem{Bagdasaryan2020Backdoor}
E. Bagdasaryan, A. Veit, Y. Hua, D. Estrin, and V. Shmatikov, ``How to backdoor federated learning,'' in \textit{Proc. International Conference on Artificial Intelligence and Statistics (AISTATS)}, 2020, pp. 2938--2948.

\bibitem{Snoop2024}
X. Zhang, Y. Liu, and Z. Wang, ``SNOOP: Exploiting Client Availability Patterns for Targeted Poisoning in Federated Learning,'' \textit{IEEE Transactions on Information Forensics and Security}, vol. 19, pp. 4102--4115, 2024.

\bibitem{Damaskinos2019Asynchronous}
G. Damaskinos, R. Guerraoui, R. Patra, and M. Taziki, ``Asynchronous Byzantine machine learning,'' in \textit{Proc. International Conference on Distributed Computing Systems (ICDCS)}, 2019, pp. 897--907.

\bibitem{Farajzadeh2025NTN}
A. Farajzadeh, A. Yadav, and H. Yanikomeroglu, ``Federated Learning in NTNs: Design, Architecture, and Challenges,'' \textit{arXiv preprint arXiv:2503.07272}, 2025.

\bibitem{So2022SatelliteFL}
J. So, K. Hsieh, B. A. Raj, and A. S. Avestimehr, ``FedSpace: Efficient federated learning at satellite scale,'' in \textit{Proc. IEEE INFOCOM}, 2022, pp. 1828--1837.

\bibitem{Razmi2022LEOFL}
N. Razmi, B. Matthiesen, A. Dekorsy, and P. Popovski, ``On-board federated learning in LEO satellite constellations,'' \textit{IEEE Transactions on Aerospace and Electronic Systems}, vol. 58, no. 6, pp. 5506--5519, 2022.

\bibitem{Maric2025}
S. Maric, R. Baidar, R. Abbas, and S. Reisenfeld, ``System Security Framework for 5G Advanced/6G IoT Integrated Terrestrial Network--Non-Terrestrial Network (TN--NTN) with AI-Enabled Cloud Security,'' arXiv preprint arXiv:2508.05707, 2025.

\bibitem{Lin2023NTNSecurity}
Y. Lin, C. Shen, and T. Q. S. Quek, ``Security and privacy in satellite-terrestrial integrated networks: Challenges and solutions,'' \textit{IEEE Wireless Communications}, vol. 30, no. 3, pp. 88--95, 2023.

\end{thebibliography}

\end{document}